\documentclass{article}

\usepackage{arxiv}

\usepackage[utf8]{inputenc} % allow utf-8 input
\usepackage[T1]{fontenc}    % use 8-bit T1 fonts
\usepackage{hyperref}       % hyperlinks
\usepackage{url}            % simple URL typesetting
\usepackage{booktabs}       % professional-quality tables
\usepackage{amsfonts}       % blackboard math symbols
\usepackage{nicefrac}       % compact symbols for 1/2, etc.
\usepackage{microtype}      % microtypography
\usepackage{lipsum}		% Can be removed after putting your text content
\usepackage{graphicx}
\usepackage{natbib}
\usepackage{doi}
\usepackage{amsmath}
\usepackage{subcaption}
\usepackage{multirow}
\usepackage{xcolor}
\usepackage[draft]{changes}

\title{DETNO: A Diffusion-Enhanced Transformer Neural Operator for Long-Term Traffic Forecasting}

\author{ {\hspace{1mm}Owais Ahmad} \\
	Applied Research, Quantiphi\\
	Marlborough, MA 01752, USA\\
	\texttt{owais.ahmad@quantiphi.com} \\
	\And
	{\hspace{1mm}Milad Ramezankhani} \\
	Applied Research, Quantiphi\\
	Marlborough, MA 01752, USA\\
	\texttt{milad.ramezankhani@quantiphi.com} \\
    \And
	{\hspace{1mm}Anirudh Deodhar} \\
	Applied Research, Quantiphi\\
	Marlborough, MA 01752, USA\\
	\texttt{anirudh.deodhar@quantiphi.com} \\
}

\renewcommand{\shorttitle}{\textit{arXiv} Template}

\hypersetup{
pdftitle={A template for the arxiv style},
pdfsubject={q-bio.NC, q-bio.QM},
pdfauthor={David S.~Hippocampus, Elias D.~Striatum},
pdfkeywords={First keyword, Second keyword, More},
}

\begin{document}
\maketitle

\begin{abstract}

Accurate long-term traffic forecasting remains a critical challenge in intelligent transportation systems, particularly when predicting high-frequency traffic phenomena such as shock waves and congestion boundaries over extended rollout horizons. Neural operators have recently gained attention as promising tools for modeling traffic flow. While effective at learning function space mappings, they inherently produce smooth predictions that fail to reconstruct high-frequency features such as sharp density gradients which results in rapid error accumulation during multi-step rollout predictions essential for real-time traffic management. To address these fundamental limitations, we introduce a unified Diffusion-Enhanced Transformer Neural Operator (DETNO) architecture. DETNO leverages a transformer neural operator with cross-attention mechanisms, providing model expressivity and super-resolution, coupled with a diffusion-based refinement component that iteratively reconstructs high-frequency traffic details through progressive denoising. This overcomes the inherent smoothing limitations and rollout instability of standard neural operators. Through comprehensive evaluation on chaotic traffic datasets, our method demonstrates superior performance in extended rollout predictions compared to traditional and transformer-based neural operators, preserving high-frequency components and improving stability over long prediction horizons. 

% This enables accurate reconstruction of sharp traffic transitions and localized phenomena. DETNO offers a robust solution for real-time traffic density and velocity estimation from sparse stationary sensor observations, providing direct applications in  adaptive traffic control systems that demand reliable long-term predictions.

\end{abstract}

% keywords can be removed
% \keywords{Traffic Forecasting. Transformer. Diffusion Models. Neural Operator}

\section{Introduction}

Precise traffic forecasting is essential for effective transportation system management, particularly as urbanization accelerates and mobility needs continue to grow~\cite{lana2018road}. 
% With over 68\% of the global population expected to live in urban areas by 2050, there is an urgent need for scalable, deployable traffic management solutions that can be rapidly implemented across diverse metropolitan areas~\cite{undesa2018urbanization}. The increasing complexity of urban traffic networks has intensified the demand for sophisticated traffic estimation methodologies that can provide real-time insights into traffic dynamics. 
Traffic modeling, however, faces significant challenges due to sparse data availability and the chaotic, nonlinear dynamics of traffic flow that include sudden transitions, congestion formation, and shockwave propagation, making accurate prediction exceptionally difficult~\cite{alghamdi2022comparative,smith1997traffic}. 
% The nonlinear interactions between vehicles, varying driver behaviors, and environmental factors create intricate patterns that challenge traditional modeling approaches. 
Conventional traffic data collection methods (e.g., loop detectors, cameras and probe vehicles) offer valuable insights within their coverage areas. However, real-world data has inherent limitations for inverse problems and optimization tasks that require controlled conditions, systematic parameter variation, or counterfactual analysis~\cite{jain2012road,shafik2025real}. Many critical traffic engineering problems therefore rely on using numerical solvers to simulate traffic flow and generate synthetic data, providing the reproducible environments needed for advanced analysis.
% One alternative approach is to use numerical solvers to model traffic flow and generate synthetic data for modeling purposes. To collect data, simulation approaches rely on partial differential equations like the Lighthill-Whitham-Richards (LWR) model, which describes traffic dynamics through vehicle conservation laws~\cite{lighthill1955kinematic}. While the LWR model effectively captures phenomena such as shockwave propagation and congestion formation, solving it requires conventional numerical methods like the Godunov scheme, which demand careful discretization and stability considerations, introducing substantial computational overhead. Second-order models, including the Aw-Rascle-Zhang (ARZ) system~\cite{aw2000resurrection}, address non-equilibrium conditions but further amplify computational challenges. This computational limitation prevents deployment of such methods in city-wide traffic management centers where real-time decision making is crucial for managing thousands of intersections and highway segments simultaneously.
Models based on partial differential equations, such as the Lighthill-Whitham-Richards (LWR) model~\cite{lighthill1955kinematic}, capture key dynamics like shockwave propagation and congestion, but require computationally intensive schemes (e.g., Godunov) with strict discretization and stability constraints. Second-order models like Aw-Rascle-Zhang (ARZ)~\cite{aw2000resurrection} handle non-equilibrium conditions but further increase complexity, limiting their use for real-time city-wide traffic management.

To address these computational and data sparsity challenges, machine learning (ML) models have emerged as a promising solution. These approaches can learn complex traffic patterns from available data while being more computationally efficient than traditional numerical solvers for real-time applications. Importantly, ML-based solutions offer significant advantages for large-scale deployment, including the ability to process multiple traffic scenarios in parallel and adapt to varying urban infrastructure without requiring extensive recalibration for each new deployment site.
\begin{figure*}[ht]
    \centering
    \includegraphics[width=\textwidth]{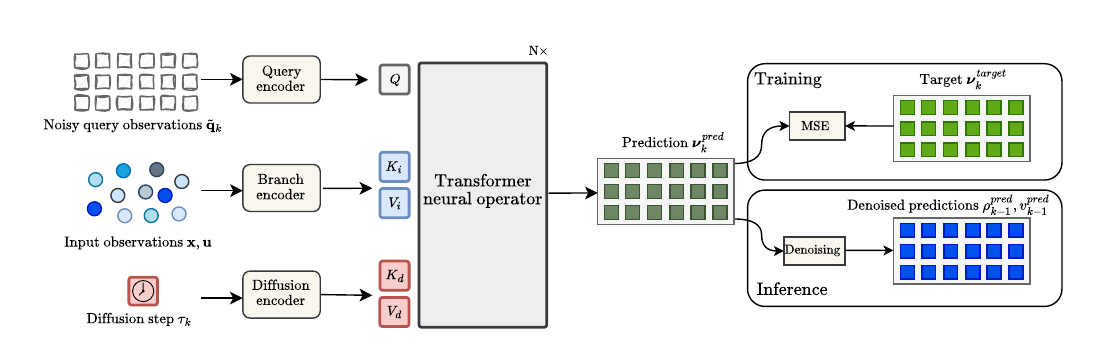}
    \caption{DETNO architecture for traffic forecasting. Noisy query tokens $\tilde{\mathbf{q}}_{k}=[x,t,\tilde{\rho}_{k},\tilde{v}_{k}]$ are processed by the \emph{Query encoder}; sensor observations $[\mathbf{x},\mathbf{u}]$ (coordinates and measurements) are processed by the \emph{Branch encoder}; and the diffusion timestep $\tau_{k}$ (Fourier-embedded) is processed by the \emph{Diffusion encoder}. Their outputs form a query stream $\mathbf{Q}$ and two context streams, operator stream $(\mathbf{K}_{i},\mathbf{V}_{i})$ and diffusion stream $(\mathbf{K}_{d},\mathbf{V}_{d})$, which a transformer neural operator processes via heterogeneous cross-attention (followed by self-attention). At step $k$, the model predicts the diffusion velocity $\boldsymbol{\nu}^{pred}_{k}$. During training, a $v$-parameterization loss minimizes MSE to the target $\boldsymbol{\nu}_{k}^{target}$; during inference, $\boldsymbol{\nu}^{pred}_{k}$ drives a DDIM update to recover $(\rho^{pred}_{k-1},v^{pred}_{k-1})$ in a $k\!\to\!k\!-\!1$ denoising schedule, yielding the final predictions.}
    \label{fig:pipeline}
\end{figure*}
ML-based approaches such as Graph Neural Networks (GNNs) model traffic networks as graphs, with nodes as road segments and edges as connectivity~\cite{yu2017spatio,peng2020spatial}. Advanced variants like Graph Attention Networks (GATs)~\cite{zhang2019spatial,kong2020stgat,wang2022attention} have demonstrated superior performance in capturing non-linear spatial correlations and temporal dynamics through attention mechanisms. However, these models heavily rely on data availability and can suffer from poor generalization in new traffic scenarios. Scientific ML approaches such as Physics-Informed Neural Networks (PINNs) \cite{raissi2019physics} have recently shown significant promise in learning traffic flow dynamics ~\cite{shi2021physics,usama2022physics}. They embed traffic flow physics into the learning process, enabling robust data-agnostic modeling. However, their poor domain generalization in new initial/boundary conditions and high computational cost for enforcing physics constraints hinder deployment in real-time, city-scale traffic management.
Neural operators have been introduced to address the generalization limitations of PINNs \cite{lu2019deeponet, li2020fourier}. Models such as DeepONets~\cite{rap2025traffic}, which learn mappings between function spaces, Fourier Neural Operators (FNOs)~\cite{thodi2024fourier}, which capture traffic dependencies in the frequency domain, and Variable-Input Deep Operator Networks (VIDON)~\cite{prasthofer2022variable}, which handle irregular and heterogeneous network layouts, have been successfully applied to traffic forecasting. These methods generalize across diverse road conditions, sensor configurations, and network densities without retraining by learning fundamental traffic flow governing laws. However, neural operators suffer from spectral bias \cite{moseley2023finite, ramezankhani2025advanced}, favoring low-frequency components and producing overly smooth predictions that fail to capture high-frequency phenomena such as shockwaves, and abrupt congestion transitions critical for effective traffic management. This limitation can be detrimental in long temporal rollouts, leading to error accumulation and significance divergence from ground truth.

To mitigate the spectral bias of neural operators, two diffusion-based families have emerged: two-stage and single-stage strategies. The former trains a neural operator first, and then a score-based diffusion model is conditioned on the neural operator's outputs to restore high-frequency detail and improve spectral alignment~\cite{oommen2025integratingneuraloperatorsdiffusion,perrone2025integrating,guo2025physics}.  While it is able to match the ground-truth spectrum, its added fine-scale detail can be partially \textit{hallucinated} and \textit{non-physical}, leading to limited gains in pointwise errors like mean squared error (MSE) and extra inference cost from multiple denoising steps. The second appoach leverages diffusion-inspired multistep denoising  within the neural operator to iteratively reweight non-dominant (i.e., high-frequency) components and improve long-horizon stability~\cite{lippe2023pde, serrano2024aroma}.  However, current realizations typically rely on conventional backbones (e.g., UNet and FNO) operating on fixed and regular grids, which can constrain long-range and multi-scale spatiotemporal modeling and limit portability across real-world heterogeneous geometries.

To address the aforementioned limitations, we introduce the Diffusion-Enhanced Transformer Neural Operator (DETNO), an end-to-end architecture that couples a transformer neural operator with a diffusion refiner \emph{within one unified model}. DETNO leverages a heterogeneous cross-attention module that maintains two distinct information streams: (i) an \emph{operator stream} whose keys/values encode input functions (e.g., sensor fields and boundary/initial conditions), and (ii) a \emph{diffusion stream} whose keys/values encode the diffusion noise level/timestep.  Each query derived from spatiotemporal coordinates is \emph{conditioned by} both K/V sets, ensuring both the input functions and diffusion schedule influence the predictions in a controlled manner. The DETNO architecture also enables super-resolution, allowing to query at any arbiterory resolution in the traffic spatiotemporal domain. The integrated diffusion refiner performs a small number of iterative denoising steps to reconstruct fine-grained structure, recovering high-frequency phenomena with a modest computational overhead. The main contributions of this paper are twofold: (1) we introduce a novel diffusion-enhanced neural operator architecture that explicitly addresses error accumulation over long temporal rollouts in traffic forecasting; and (2) we demonstrate that the method significantly outperforms neural operator baselines by effectively capturing high-frequency traffic dynamics and sharp transitions that conventional approaches typically smooth out.

\section{Methodology}

% \subsection{DETNO Architecture}
% Figure~\ref{fig:pipeline} illustrates our DETNO pipeline, which unifies transformer-based neural operators with diffusion refinement in a single end-to-end architecture. The model processes sparse sensor measurements from fixed highway locations and generates refined traffic predictions at arbitrary query positions.
% The architecture operates through an integrated processing pipeline where sensor measurements and query locations are encoded separately before being combined through cross-attention mechanisms. The transformer component establishes global traffic context by learning relationships between sensor observations and prediction targets, while the mixture-of-experts(MoE) framework enables specialized processing for different traffic scenarios and spatial regions.
% Unlike traditional approaches that treat coarse prediction and refinement as separate stages, DETNO jointly optimizes both components within a unified objective function. The transformer neural operator provides spatially-aware representations that serve as effective conditioning for the diffusion process, while the diffusion mechanism guides the transformer to learn representations that are conducive to high-quality refinement.

\subsection{Transformer Neural Operator}

As illustrated in Figure \ref{fig:pipeline}, our transformer neural operator comprises a heterogeneous cross-attention block followed by self-attention. The cross-attention exposes two context streams as keys/values: an operator stream derived from input functions and a diffusion stream derived from the diffusion timestep; queries contain spatiotemporal coordinates at which we predict traffic states. All three inputs are first mapped by dedicated encoders: a Query encoder, a Branch encoder (operator stream), and a Diffusion encoder. Each encoder is an multi-layer perceptron (MLP) that projects its input into a $d$-dimensional latent; the Diffusion encoder additionally applies a Fourier embedding to the timestep before the MLP. For each query, we compute linear cross-attention \cite{hao2023gnot} separately against the operator and diffusion streams to produce two context vectors that are then fused (summation and projection with a residual) into an updated query representation; a subsequent self-attention layer lets queries exchange information and enforce spatial–temporal coherence. We use a Mixture-of-Experts (MoEs) in the transformer blocks, with a gating network conditioned on each query’s spatiotemporal coordinates to realize a soft domain decomposition that routes tokens to specialized experts \cite{hao2023gnot, ramezankhani2025gito}.

The training data consist of two sets with different cardinalities: sensors $\{(\mathbf{x}^{i},\mathbf{u}^{i})\}_{i=1}^{N_{\text{sensor}}}$ and queries $\{(\mathbf{q}^{j},\mathbf{y}^{j})\}_{j=1}^{N_{\text{pred}}}$, where $\mathbf{x}^{i}\in\mathbb{R}^{2}$ are space–time coordinates $(x,t)$, $\mathbf{u}^{i}\in\mathbb{R}^{2}$ are traffic sensor states $(\rho,v)$, $\mathbf{q}^{j}\in\mathbb{R}^{4}$ are query tokens $[x_q,t_q,\rho,v]$, and $\mathbf{y}^{j}\in\mathbb{R}^{2}$ are ground-truth states at the same query locations. The encoders map inputs to width $d$ as follows: the Query encoder $\phi_q:\mathbb{R}^{4}\!\to\!\mathbb{R}^{d}$ applies an MLP to yield $\mathbf{Q}\in\mathbb{R}^{N_{\text{pred}}\times d}$; the Branch encoder $\phi_b:\mathbb{R}^{4}\!\to\!\mathbb{R}^{d}$ (MLP) acts on $[\mathbf{x}^{i},\mathbf{u}^{i}]$ to produce operator-stream keys/values $(\mathbf{K}_{i},\mathbf{V}_{i})\in\mathbb{R}^{N_{\text{sensor}}\times d}$; the Diffusion encoder maps the diffusion timestep $\tau\in\mathbb{R}$ through a Fourier embedding $\gamma(\tau)\in\mathbb{R}^{d_\tau}$ and an MLP to $z_{d}\in\mathbb{R}^{d}$, which is broadcast across queries to form $(\mathbf{K}_{d},\mathbf{V}_{d})\in\mathbb{R}^{N_{\text{sensor}}\times d}$. Linear cross-attention is computed separately against the operator and diffusion streams,
% \[
\begin{equation}
\mathbf{C}_{i}=\mathrm{Attn}(\mathbf{Q},\mathbf{K}_{i},\mathbf{V}_{i}),\ \ \ 
\mathbf{C}_{d}=\mathrm{Attn}(\mathbf{Q},\mathbf{K}_{d},\mathbf{V}_{d}).
\end{equation}
% \]
The outputs are then fused and passed through linear self-attention and a feed-forward block to produce a diffusion velocity fields $\hat{\boldsymbol{\nu}}\in\mathbb{R}^{N_{\text{pred}}\times 2}$ (for both traffic density and velocity) at the query coordinates. The composition of $\mathbf{q}^{j}$ differs between training and inference: in training, its $(\rho,v)$ entries are noise-corrupted versions of $\mathbf{y}^{j}$; at inference, they are initialized with pure noise and refined by the diffusion process, allowing a single query format for both supervised learning and test-time denoising.

\subsection{Diffusion-Based Refinement}
The diffusion refiner learns to remove injected noise from traffic states at query locations using a velocity-parameterization objective. During training, for a noise level $k\in\{0,\dots,K\}$, we draw $\boldsymbol{\epsilon}\sim\mathcal{N}(0,\mathbf{I})$ and form the corrupted targets
$\tilde{\mathbf{y}}_{k}=\sqrt{\bar{\alpha}_{k}}\,\mathbf{y}+\sqrt{1-\bar{\alpha}_{k}}\,\boldsymbol{\epsilon}$,
where $\mathbf{y}$ is the clean state (density, velocity) at the query locations and $\bar{\alpha}_{k}$ is the cumulative noise schedule ~\cite{song2020denoising}. The noisy query tokens are generated as
$\tilde{\mathbf{q}}_{k}=[x_q,t_q,\tilde{\rho}_{k},\tilde{v}_{k}]$, with $(\tilde{\rho}_{k},\tilde{v}_{k})$ taken \emph{directly} from the two channels of $\tilde{\mathbf{y}}_{k}$ at the same query coordinates. Conditioned on $(\mathbf{x},\mathbf{u},\tilde{\mathbf{q}}_{k},\tau_k)$, the DETNO model $\mathcal{G}_{\theta}$ predicts the diffusion velocity $\boldsymbol{\nu}_{k}$. The supervision target uses the standard $v$-parameterization
\begin{equation}
\boldsymbol{\nu}_{k}^{*}=\sqrt{\bar{\alpha}_{k}}\,\boldsymbol{\epsilon}-\sqrt{1-\bar{\alpha}_{k}}\,\mathbf{y},
\end{equation}
and the diffusion loss for a single diffusion process is
\begin{equation}
\mathcal{L}_{\text{diffusion}}=\sum_{k=0}^{K}\mathbb{E}_{k,\boldsymbol{\epsilon}}\left\|\mathcal{G}_{\theta}(\mathbf{x},\mathbf{u},\tilde{\mathbf{q}}_{k},\tau_k)-\boldsymbol{\nu}_{k}^{*}\right\|_2^{2}.
\end{equation}

At inference, we initialize the traffic state entries with pure noise (i.e., $\rho_K$ and $v_K$), $\mathbf{q}^{(K)}=[x_q,t_q,\rho_K,v_K]$, and iteratively denoise over $k=K,\dots,0$. At step $k$, the model consumes $(\mathbf{x},\mathbf{u},\tilde{\mathbf{q}}_{k},\tau_k)$, outputs $\hat{\boldsymbol{\nu}}^{(k)}$, and a DDIM update produces $\rho^{pred}_{k-1}$ and $v^{pred}_{k-1}$. DDIM is preferred over DDPM for its deterministic updates that permit larger step sizes and fewer evaluations while preserving quality, yielding faster denoising without changing the training loss~\cite{song2020denoising}. It is crucial to distinguish the physical time $t_q$ in the query coordinates from the denoising timestep $\tau$ used by the diffusion stream. This timestep tells the model how much noise is present in the current prediction, acting as a progress indicator during the refinement process. At each refinement step, the current diffusion timestep is computed as $\tau_k=\mathrm{scheduler\_timestep}(k)\cdot\frac{1000}{K}$. The resulting embedding $\gamma(\tau_k)$ (Fourier features followed by an MLP, as defined previously) is used to form the diffusion-stream keys/values $\mathbf{K}_d,\mathbf{V}_d$.

\subsection{Case Study: Long-term Traffic Forecasting}
To evaluate the proposed DETNO approach for traffic prediction, we establish a controlled simulation environment that models traffic flow dynamics over a spatiotemporal domain. The setup represents a highway segment with fixed sensors providing sparse observations of density and velocity at discrete spatiotemporal locations, mimicking real-world monitoring. The goal is to predict traffic states $\boldsymbol{u}(x,t)=[\rho(x,t),v(x,t)]^\top$ over $X:=[x_{\min},x_{\max}]\subset\mathbb{R}$ and $T\subset\mathbb{R}^+$, with current time $t_c\in T$ and windows $\Delta_{\text{past}},\Delta_{\text{pred}}>0$ such that $t_c-\Delta_{\text{past}},\,t_c+\Delta_{\text{pred}}\in T$. Traffic flow dynamics are governed by the Lighthill-Whitham-Richards (LWR) model \cite{lighthill1955kinematic} describing traffic density evolution through vehicle conservation:
\begin{equation}
\frac{\partial \rho(x, t)}{\partial t} + \frac{\partial}{\partial x} \left[\rho(x, t) \cdot v(\rho(x, t))\right] = 0, \quad (x, t) \in X \times T
\end{equation}
where $v(\rho)$ represents the velocity-density relationship and $\rho(x, t)$ denotes the traffic density. We formulate the traffic estimation problem as learning an operator $\mathcal{G}$ from the input function space $\mathcal{A}$ to the solution space $\mathcal{H}$ over domain $X \times T$, i.e., $\mathcal{G} : \mathcal{A} \rightarrow \mathcal{H}$, where $\mathcal{H}$ is the function space over the spatiotemporal domain. The input function space $\mathcal{A}$ contains sensor measurements, spatiotemporal coordinates and boundary condition data from road endpoints. For any input configuration at current time $t_c$, the operator maps stationary sensor observations combined with boundary conditions to complete traffic state fields $\boldsymbol{u}(x, t)$ at arbitrary spatiotemporal query locations within the prediction time window.
% \deleted{For data generation, traffic density evolves according to the Lighthill–Whitham–Richards (LWR) model,
% \begin{equation}
% \frac{\partial \rho(x,t)}{\partial t}+\frac{\partial}{\partial x}\big[\rho(x,t)\,v(\rho(x,t))\big]=0,\quad (x,t)\in X\times T,
% \end{equation}
% where $\rho$ is the density field and $v(\rho)$ is the fundamental diagram (velocity–density relation). Synthetic trajectories used for training and evaluation are produced with a first-order Godunov finite-volume scheme. Implementation details are provided in the Supplementary Information.}

\begin{figure}[h]
    \centering
    \includegraphics[width=0.8\textwidth]{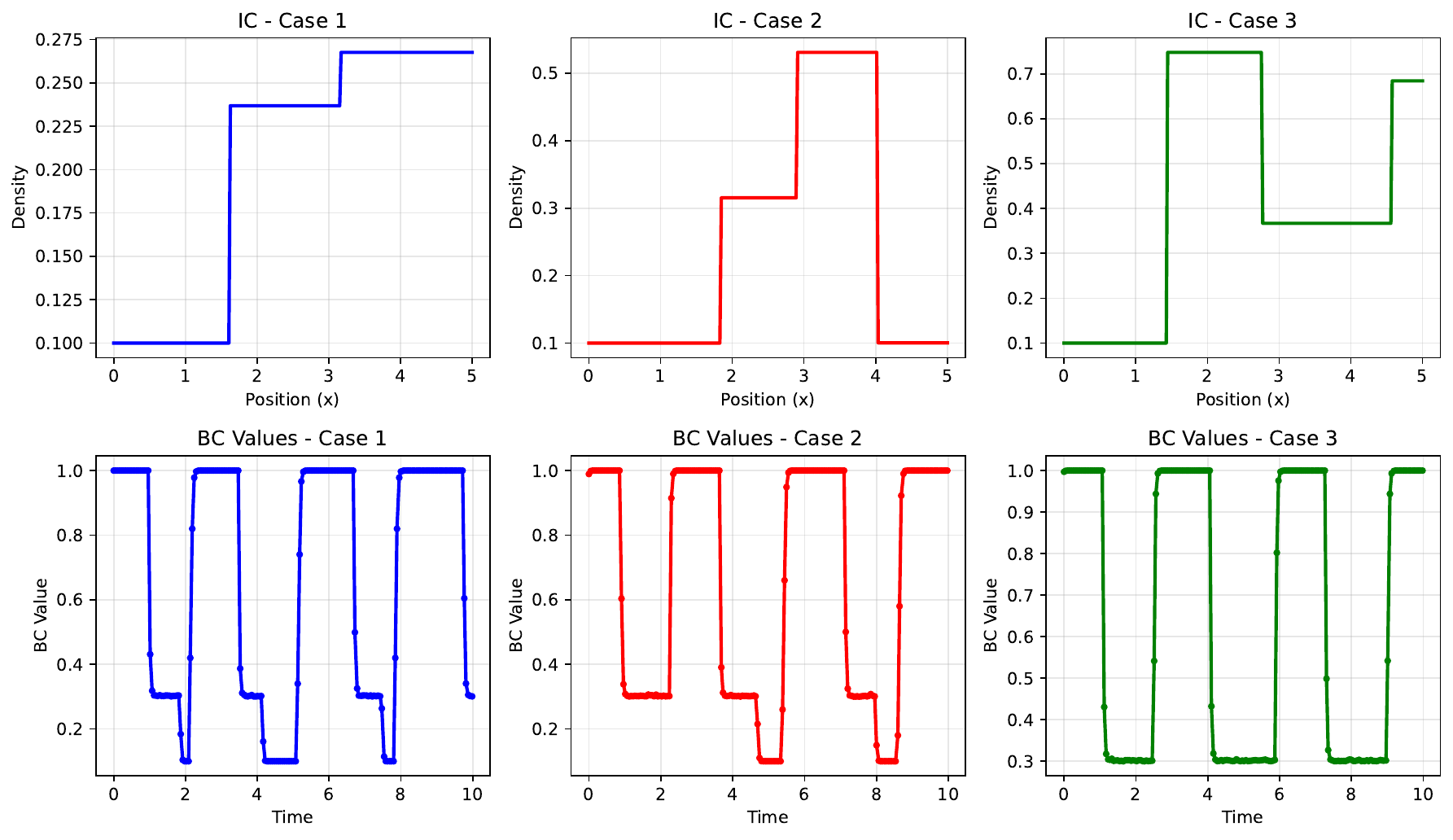}
    \caption{Chaotic initial and boundary conditions for the Godunov dataset. The multi-step piecewise constant density profiles (top row) and time-varying boundary conditions simulating traffic light cycles (bottom row) generate complex spatiotemporal traffic dynamics with sharp density transitions and discontinuous patterns that challenge standard neural operator approaches.}
    \label{fig:godunov}
\end{figure}

We generate synthetic traffic data using the Godunov numerical scheme \cite{friedrich2018godunov} to solve the LWR traffic flow equation on a $5$ kilometer highway segment for a total traffic simulation time of 25 min. Initial conditions are created using multi-step piecewise constant density profiles with randomly sampled parameters, where the initial density $\rho(x, 0)$ is constructed by adding random step functions to a base inflow density of $\rho_{\text{init}} = 0.1$. Boundary conditions simulate traffic light control at the downstream end, with alternating red and green phases lasting $1$-$2$ minutes each, creating realistic congestion patterns. The dataset uses sensor measurements from fixed locations along the highway with a history window of $\Delta_{\text{past}} = 1$ minute and prediction horizon of $\Delta_{\text{pred}} = 1$ minute for single-step training. The initial and boundary conditions are generated by assigning random step values to the density, as illustrated in Figure~\ref{fig:godunov}.The boundary conditions are combined with interior sensor measurements in each training sample, giving the model both in-domain traffic observations and terminal state information. 
During rollout, DETNO processes sensor data and boundary conditions jointly within each time window. 
% For predictions over $[t_c + \Delta t, t_c + 2\Delta t]$, boundary data is available across the extended domain $[t_c, t_c + 2\Delta t]$, ensuring coverage of both the historical context and the target period.

\section{Results and Discussion}

\subsection{Experiment settings}

\subsubsection{DETNO Architecture}
In this section, we provide a detailed description of the model architecture, outlining the design and functionality of each component block.

% \begin{itemize}
\textbf{Sensor Encoding Block.} Input sensor coordinates $(x, t)$ and traffic measurements (density, velocity) are concatenated and processed through a $2$-layer MLP with hidden dimension $64$ and GELU activation to produce sensor representations.

\textbf{Query Token Encoding Block.} Query tokens containing spatial-temporal coordinates $(x, t)$ and traffic state values (density, velocity) are processed through a $2$-layer MLP with hidden dimension $64$ and GELU activation to produce query embeddings.

\textbf{Time Embedding Block.} Timestep values are converted to Fourier embeddings using sine and cosine functions across $64$ frequencies with maximum period $10000$, then processed through a $2$-layer MLP (dimensions $64 \rightarrow 64 \rightarrow 64$) with GELU activation.

\textbf{Cross-Attention Block ($\times 3$ layers).} Each block contains:
\begin{enumerate}
    \item Linear cross-attention with $4$ heads processing query tokens against sensor representations and time embeddings as two input streams.
    \item MoE with $3$ expert networks, each being a $2$-layer MLP ($64 \rightarrow 256 \rightarrow 64$) with GELU activation.
    \item Spatiotemporal gating network ($4 \rightarrow 256 \rightarrow 256 \rightarrow 4$) for expert routing based on full query token dimensions.
    \item Linear self-attention with $4$ heads for spatiotemporal dependency modeling.
\end{enumerate}

\textbf{Output Projection Block.} Final representations are processed through a $2$-layer MLP with hidden dimension $64$ and GELU activation to produce traffic state predictions (density, velocity).

\textbf{Diffusion Wrapper.} The diffusion mechanism wraps the entire transformer neural operator within a denoising framework using DDIM scheduler. The wrapper manages noise scheduling through $K=10$ refinement steps with trained beta values computed as $[\text{min\_noise\_std}^{k/K}]$ for $k$ in reverse order from $K$ to $0$, where min\_noise\_std $= 9 \times 10^{-2}$.

% \textit{Training Noise processing}: Takes batch input containing sensor coordinates, sensor values, and noisy query tokens constructed from output coordinates concatenated with noise-corrupted traffic states. Processes timestep through time multiplier $1000/K$ to create timestep embeddings. Feeds sensor representations, noisy query tokens, and timestep embeddings through the complete transformer neural operator pipeline (sensor encoding → query encoding → time embedding → cross-attention layers → output projection). Returns v-predictions (dimension $N_{\text{pred}} \times 2$) representing directions toward clean traffic states for single-step loss computation.

% \textit{Inference Noise processing}: Initializes random Gaussian noise (dimension $N_{\text{pred}} \times 2$), Iteratively processes through $K$ denoising steps where each iteration: (1) constructs query tokens by concatenating output coordinates with current noisy predictions, (2) creates timestep embeddings by multiplying current scheduler timestep by time multiplier $1000$, (3) processes complete input (sensor data, query tokens, timestep) through transformer neural operator to obtain v-predictions, (4) applies DDIM scheduler step function to v-predictions and current noisy state and finally Returns final clean traffic state predictions (dimension $N_{\text{pred}} \times 2$) after complete denoising sequence.
% \end{itemize}

\subsubsection{Training Procedure and Temporal Rollout}
DETNO is trained using a \emph{single-step} prediction framework on the Godunov dataset introduced earlier. It learns a supervised mapping from sparse sensor measurements in a past window $[t_c-\Delta_{\text{past}},\,t_c]$ to traffic states at arbitrary query locations within the immediate future window $[t_c,\,t_c+\Delta_{\text{pred}}]$. In this work, both $\Delta_{\text{past}}$ and $\Delta_{\text{pred}}$ are set to 1 minute. During training, ground-truth fields are available over the full domain, so the network learns to reconstruct complete traffic states from the past sensor history for one horizon. Long-term forecasting is performed \emph{only at inference} via an autoregressive rollout. The first prediction window uses the real sensor data. For each subsequent window, we form pseudo-sensor inputs by sampling the model’s previous predictions at the fixed sensor coordinates (shifted forward in time by $\Delta_{\text{pred}}$) and combining them with boundary-condition data; these inputs are then fed back into the model to predict the next window. Iterating this procedure extends forecasts over the desired temporal horizon. In this work, we examined the models performance for 8 rollout steps. We generated $1300$ traffic simulations by varying initial and boundary conditions (e.g., initial vechicle density and velocity, and traffic-light settings at the end of the road); $1000$ samples are used for training and $300$ for testing. The diffusion mechanism wraps the entire transformer neural operator within a denoising framework using DDIM scheduler which manages to refine details using 10 refinement steps.

\subsubsection{Baseline Models} We compare DETNO against two primary baseline approaches. ON-Traffic~\cite{rap2025traffic} utilized an advanced DeepONet architecture that directly learns mappings between sensor measurements and traffic predictions. General Neural Operator Transformer (GNOT)~\cite{hao2023gnot} processes traffic data by leveraging transformer blocks and MoEs to model complex spatio-temporal traffic patterns. These baselines allow us to evaluate the effectiveness of our unified transformer-diffusion architecture against standard neural operator approaches and assess the contribution of the diffusion refinement mechanism in capturing hig-frequency features and minimizing the error accumulation over long rollouts. 

\begin{table*}[ht]
\centering
\caption{Single step and rollout performance comparison of our proposed DETNO model against ONTraffic \cite{rap2025traffic} and GNOT \cite{hao2023gnot} on the Godunov dataset. Mean squared error (MSE) and mean absolute error (MAE) are used as comparison metrics.}
\label{tab:godunov_performance}
\begin{tabular}{@{}cccccc@{}}
\toprule
Rollout Stage & Metric & ONTraffic \cite{rap2025traffic} & GNOT \cite{hao2023gnot} & \textbf{DETNO (ours)} \\
\midrule
\multirow{2}{*}{Step 1} & Avg. MSE & 0.009 & 0.003 & 0.002 \\
& Avg. MAE & 0.038 & 0.018 & 0.022  \\
\midrule
\multirow{2}{*}{Step 8} & Avg. MSE & 0.279 & 0.019 & 0.008 \\
& Avg. MAE & 0.272 & 0.038 & 0.030 \\
\midrule
\multicolumn{2}{l}{Model Size} & 1.40M & 1.13M & 1.16M \\
\bottomrule
\end{tabular}
\end{table*}

\subsection{DETNO Performance Analysis on Chaotic Traffic Data}
Table \ref{tab:godunov_performance} compares the models performance for a single step (step 1) and rollout (step 8) predictions. To evaluate the long rollout performance, the models' predictions at the 8th rollout step are compared. Our proposed DETNO approach achieves optimal rollout performance, demonstrating a 96.0\% improvement in MSE and a 26.3\% improvement in MAE compared to GNOT, the second best model. ONTraffic exhibits significant error accumulation, with MSE and MAE increasing by 30.53$\times$ and 7.10$\times$, respectively, compared to a single-step prediction (step 1). GNOT shows improved stability, with 6.32$\times$ growth in MSE and 2.11$\times$ in MAE. In contrast, our proposed DETNO method achieves the most robust long-term performance, with only 4.39$\times$ increase in MSE and 1.34$\times$ in MAE going from step 1 to step 8 prediction. This superior rollout stability highlights DETNO’s capacity to preserve fine-scale spatiotemporal features over extended prediction horizons. While the benefits of its diffusion-based refinement mechanism are modest in the initial rollout steps, its advantage becomes increasingly pronounced over time. As errors accumulate, models that fail to capture high-frequency components and sharp density gradients (such as ONTraffic and GNOT) experience rapid performance degradation. DETNO, by contrast, effectively reconstructs these high-frequency features, maintaining temporal consistency and predictive accuracy even in complex and chaotic traffic scenarios.

\begin{figure*}[h]
    \centering
    \includegraphics[width=1\textwidth]{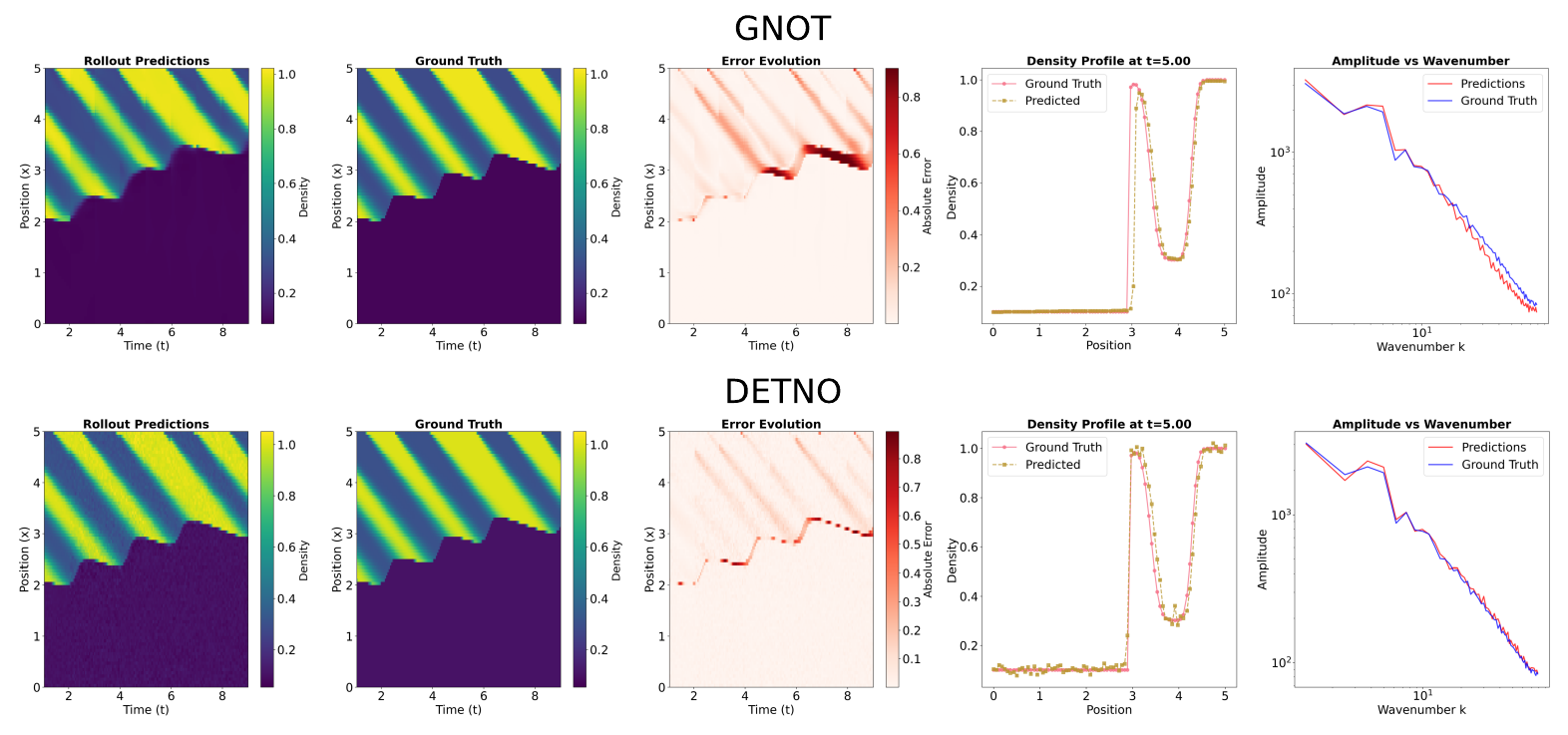}
    \caption{Comparative analysis of GNOT and DETNO predictions on chaotic traffic scenarios. The visualization shows ground truth density fields, predicted density distributions, absolute error maps, spatial density profiles at t=5.00, and frequency spectrum comparisons for two representative samples, demonstrating the refinement mechanism's superior reconstruction of sharp density transitions and localized traffic phenomena.}
    \label{fig:comparison}
\end{figure*}

To further demonstrate these rollout stability advantages, we visualized GNOT and DETNO models performance across extended prediction horizons, as illustrated in Figure \ref{fig:comparison}. 
% The analysis begins at time $t_c$ with traffic sensor measurements and boundary conditions from time window $[t_c, t_c + 2\Delta t]$ to predict traffic conditions during the overlapping window $[t_c + \Delta t, t_c + 2\Delta t]$. For the next rollout step, the model advances the time window: it treats its previous predictions from $[t_c + \Delta t, t_c + 2\Delta t]$ as additional synthetic sensors by selecting data at four key time points, then combines this synthetic sensor data with real boundary conditions and sensor measurements from the expanded window $[t_c + \Delta t, t_c + 3\Delta t]$ to predict traffic states in $[t_c + 2\Delta t, t_c + 3\Delta t]$. This process continues iteratively, creating a rolling prediction horizon that advances one minute at a time. 
The models predictions are visualized through multiple perspectives:(1) predicted density distributions, (2) ground truth density distributions, (3) absolute error heatmaps, (4) spatial density profiles at $t=5$ minutes, and (5) frequency spectrum comparisons. 
% Side-by-side heatmaps showing how predicted and actual traffic density evolve over space and time, error maps that reveal where predictions deviate from reality, cross-sectional views of traffic density at  $t=5$ minutes that highlight the model's ability to capture flow changes, and frequency analysis plots that show how well each method preserves both large-scale traffic patterns and fine-grained dynamics. These comprehensive visualizations demonstrate that while all methods perform similarly at first, DETNO's ability to refine high-frequency details becomes crucial as errors compound over time, allowing it to maintain realistic traffic patterns while other approaches gradually lose important spatial and temporal features.
 % Side-by-side heatmaps showing how predicted and actual traffic density evolve over space and time, error maps that reveal where predictions deviate from reality, cross-sectional views of traffic density at $t=5$ minutes that highlight the model's ability to capture flow changes, and frequency analysis plots that show how well each method preserves both large-scale traffic patterns and fine-grained dynamics. These comprehensive visualizations demonstrate that while all methods perform similarly at first, DETNO's ability to refine high-frequency details becomes crucial as errors compound over time, allowing it to maintain realistic traffic patterns while other approaches gradually lose important spatial and temporal features. 
The absolute-error heatmaps show that DETNO captures high-frequency structure more faithfully, yielding smaller errors along sharp transition regions (e.g., congestion fronts). In contrast, GNOT exhibits error growth over time: as the rollout progresses, both the magnitude and the spatial regions of its errors increase, consistent with missing high-frequency content early on and compounding mismatch in later steps. The spatial density profiles echo this trend: DETNO remains closely aligned with ground truth around highly nonlinear segments, whereas GNOT produces visibly smoothed transitions and local biases near discontinuities. Most critically, the frequency-spectrum comparison reveals that GNOT underestimates energy at higher wavenumbers (deviating from the ground-truth slope and amplitude), while DETNO tracks the spectrum across scales, including the high wavenumber regime. This alignment indicates that DETNO learns and preserves high-frequency features, which in turn stabilizes long-horizon rollouts and reduces error accumulation.

\subsection{Impact of Diffusion-Based Refinement}

\begin{figure}[ht]
    \centering
    \begin{subfigure}{0.48\textwidth}
        \centering
        \includegraphics[width=\linewidth]{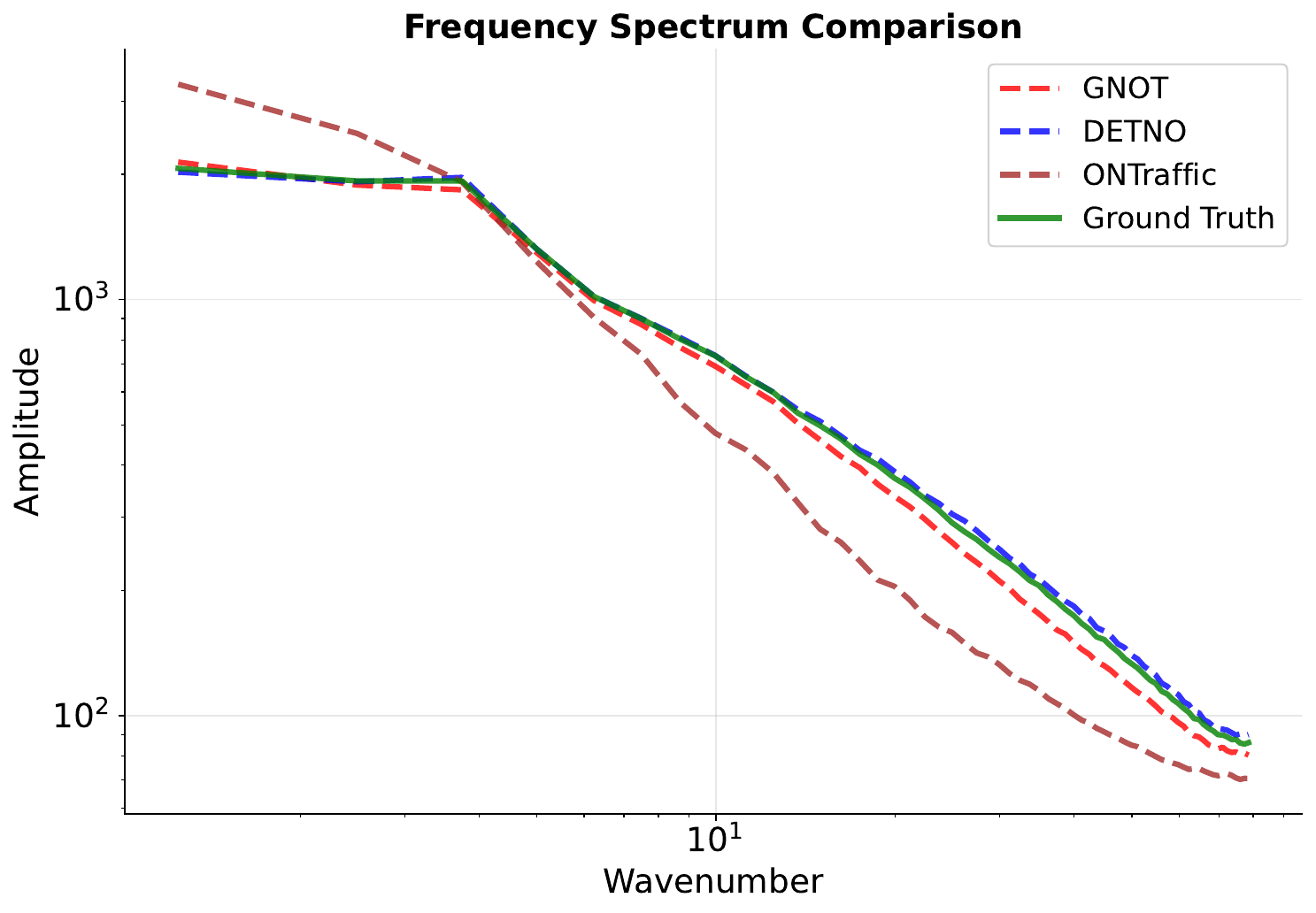}
        \caption{}
    \end{subfigure}
    \hfill
    \begin{subfigure}{0.48\textwidth}
        \centering
        \includegraphics[width=\linewidth]{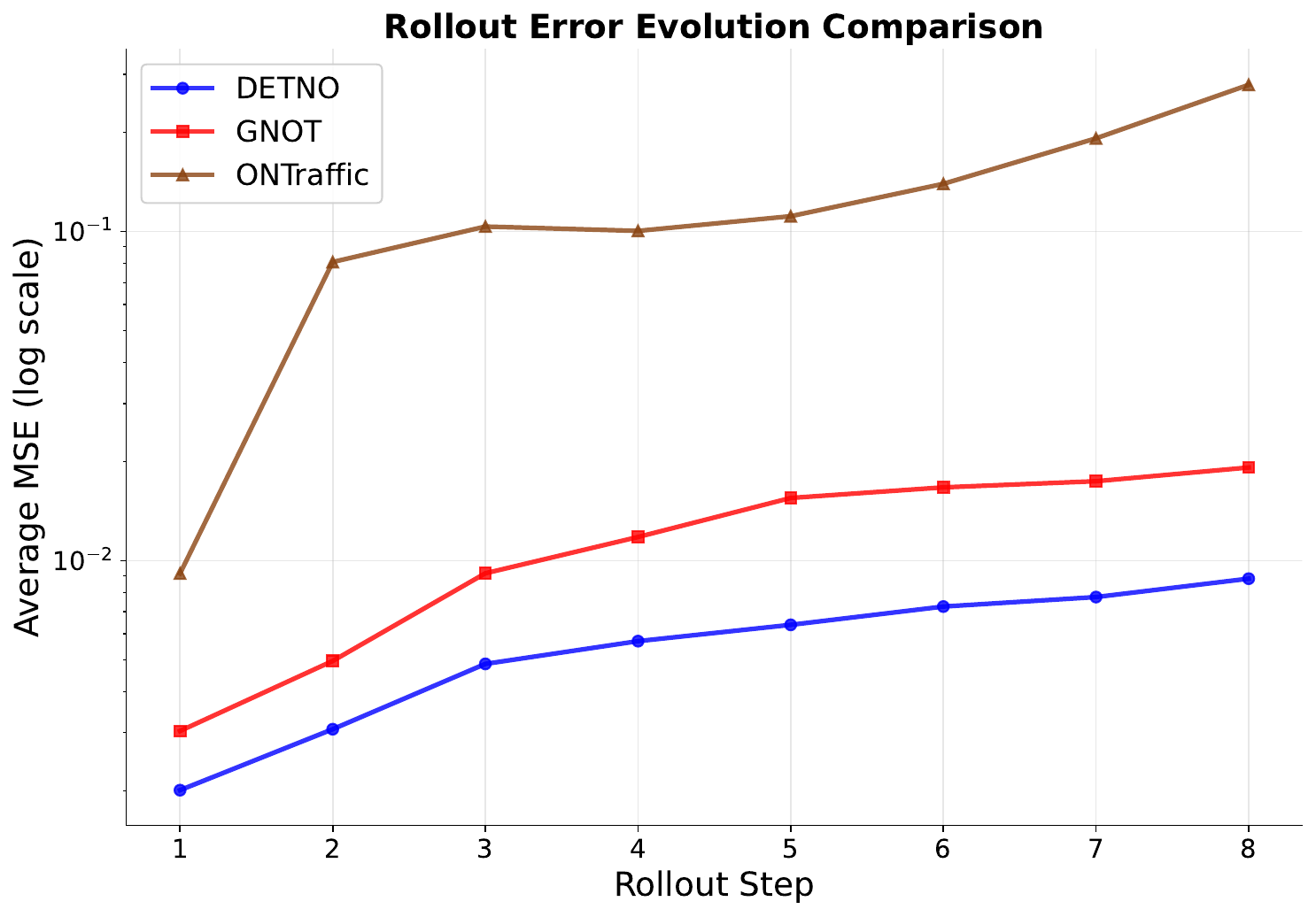}
        \caption{}
    \end{subfigure}
    \caption{Performance comparison between DETNO, GNOT and ONTraffic:
    (a) Averaged frequency spectrum across 300 samples demonstrating superior high-frequency preservation by DETNO, with enhanced amplitude retention across all wavenumbers indicating effective recovery of sharp density gradients and discontinuous traffic patterns.
    (b) Step-wise rollout error evolution showing DETNO's improved stability and accuracy over extended prediction horizons compared to GNOT and ONTraffic.}
    \label{fig:spectrum_rollout}
\end{figure}

Figure~\ref{fig:spectrum_rollout} demonstrates why DETNO sustains realistic traffic dynamics over long horizons. In panel~(a), the frequency spectra averaged over 300 test rollouts show that DETNO closely follows the ground-truth curve across scales, including the high-wavenumber regime that encodes sharp density fronts and rapidly varying congestion. By contrast, GNOT and ONTraffic exhibit a premature roll-off as wavenumber increases, indicating systematic underestimation of fine-scale energy and explaining their softened transition zones. Panel~(b) reports MSE by rollout step: errors increase for all methods with horizon length, but DETNO maintains both the lowest magnitude and the shallowest growth rate; ONTraffic is worst at all steps, and GNOT lies between ONTraffic and DETNO yet diverges more quickly than DETNO. The widening gap over time is consistent with the spectral finding: missing high-frequency content at early steps compounds under autoregressive reuse, whereas DETNO’s fidelity at high wavenumbers slows error accumulation and preserves coherent traffic patterns deeper into the rollout.

\begin{figure}[ht]
  \centering
  \includegraphics[width=0.7\textwidth]{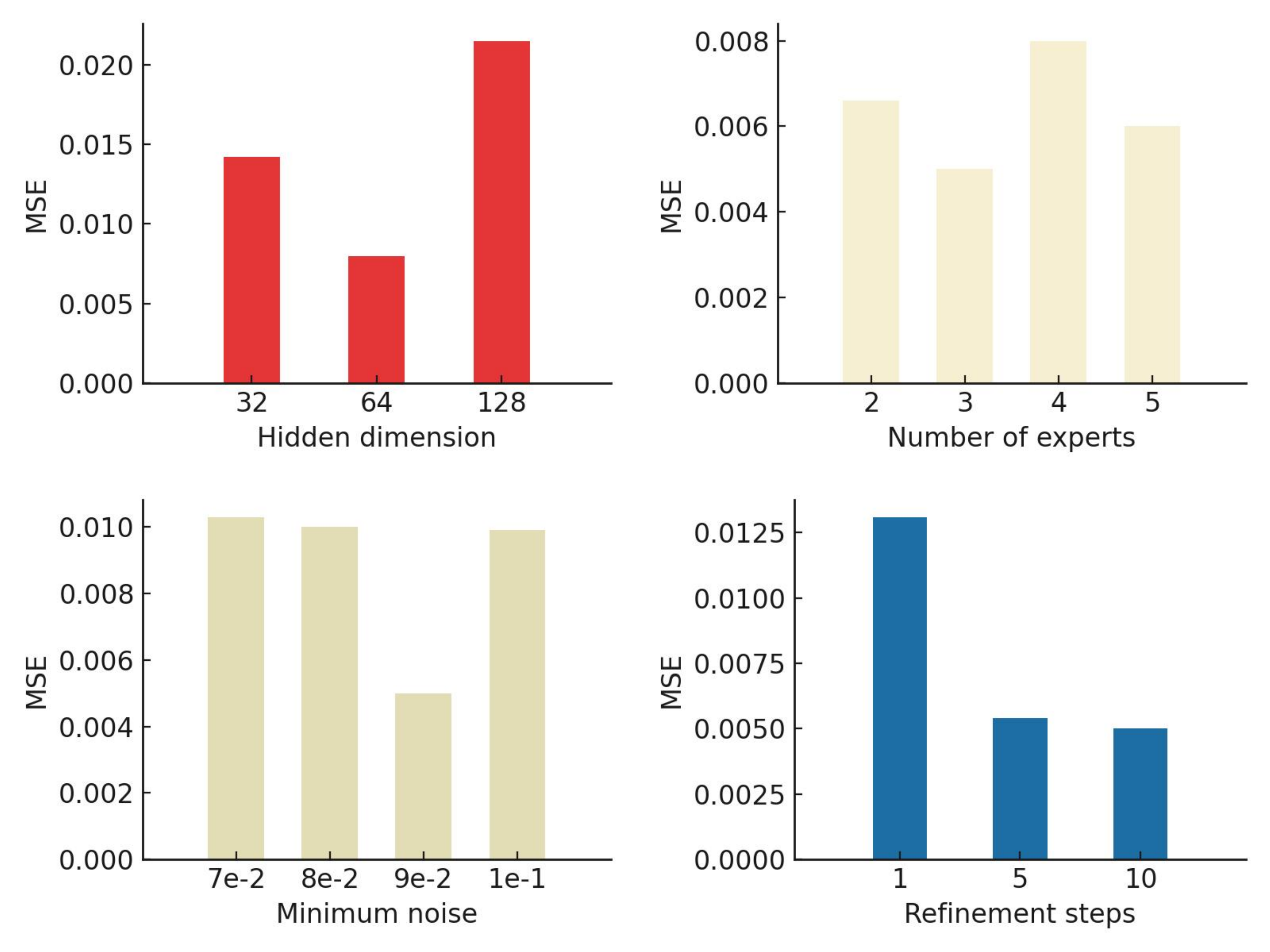}
  \caption{Ablation results for hidden dimension, number of experts, minimum noise, and refinement steps. Bars report MSE for each setting; lower is better.}
  \label{fig:ablation_mse_panel}
\end{figure}

\subsection{Ablation Studies} 
We conducted comprehensive ablation studies to determine the optimal hyperparameter setting for our proposed DETNO architecture as elaborated below (Figure \ref{fig:ablation_mse_panel}).

\textbf{Hidden dimension.} At 64 hidden units the network achieves the lowest error (MSE 0.0080), whereas reducing capacity to 32 underfits the dynamics (MSE 0.0142) and increasing to 128 degrades further (MSE 0.0215). The latter suggests over-parameterization in this data regime and less stable expert routing, leading to poorer generalization. 

\textbf{Number of experts.} Three experts provide the strongest balance between specialization and data fragmentation (MSE 0.0050). With two experts, capacity is insufficient for learning heterogeneous traffic regimes (MSE 0.0066). Adding more experts yields diminishing or negative returns: four experts markedly worsen performance (MSE 0.0080), and five experts only partially recover (MSE 0.0060). These results are consistent with MoE load-balancing effects, where increasing the expert count reduces per-expert sample density and makes routing harder to optimize \cite{fedus2022switch}. 

\textbf{Minimum noise (diffusion floor).} A “Goldilocks” level of corruption is required for effective denoising. The best setting is $9\times10^{-2}$ (MSE 0.0050). Lower noise at $7\times10^{-2}$ or $8\times10^{-2}$ weakens the learning signal and raises error (MSE 0.0103/0.0100), while a higher floor at $1\times10^{-1}$ over-corrupts targets and again increases error (MSE 0.0099). The chosen level provides enough perturbation to teach robust corrections without losing key structure. 

\textbf{Refinement steps (DDIM).} We observed that multi-step refinement is essential. A single step is inadequate (MSE 0.0131), five steps capture most of the gains (MSE 0.0054), and ten steps deliver the best overall accuracy (MSE 0.0050). The improvement from five to ten steps is modest, reflecting gradual restoration of fine-scale structure; beyond ten steps, additional latency is unlikely to be justified by further gains.

\textbf{Cross-attention design.} Using \emph{two streams} for keys/values—an operator stream for input functions and a diffusion stream for the denoising timestep—outperforms concatenating the temporal embedding with input functions into a \emph{single} stream. The two-stream design proposed in DETNO achieves the MSE of \(=0.0050\) versus \(0.0053\) for the concatenated variant. Separating streams helps the model disentangle complementary roles (sensor-driven context vs.\ denoising progress) and conditions each query on both without conflating sources, yielding more stable refinement and better fidelity at prediction query points.

Based on the above ablations, we chose the following configuarion for our DETNO architecture: 64 hidden units, 3 experts, a minimum noise of $9\times10^{-2}$,  10 refinement steps and a two-stream cross attention. This led to a better preservation of high-frequency traffic features while maintaining stable long-horizon rollouts.

% \section{Conclusions}
% This work presents a comprehensive framework for traffic forecasting that successfully addresses the fundamental limitations of existing neural operator approaches through diffusion-based refinement. The integration of transformer neural operators with diffusion mechanisms represents a significant advancement in handling sparse vehicle data while preserving critical high-frequency traffic dynamics.
% The architectural innovations establish a new paradigm for operator learning in traffic applications, where the seamless integration of global pattern recognition with local detail refinement creates a versatile framework adaptable to varying traffic complexity levels. The demonstrated capabilities position this approach where accurate, real-time traffic state estimation is essential for optimizing urban mobility and supporting autonomous vehicle deployment.

\section{Conclusion}
This work introduced DETNO, a diffusion-enhanced transformer neural operator that addresses two persistent challenges in scientific traffic forecasting: spectral bias against high-frequency features and error accumulation in long rollouts. DETNO unifies operator learning and denoising in a single stage via a heterogeneous cross-attention module that conditions queries on two distinct streams, sensor-driven input functions (operator stream) and the denoising timestep (diffusion stream), and augments capacity with a mixture-of-experts backbone and linear attention for scalability. A $v$-parameterized diffusion objective with DDIM sampling enables efficient, few-step refinement without changing the training loss, while the model’s resolution-free formulation supports super-resolution queries at arbitrary space–time coordinates. In a controlled LWR–Godunov setting, DETNO consistently outperformed neural operator baselines such as DeepONet and GNOT across qualitative and quantitative analyses.

\bibliographystyle{unsrtnat}
\bibliography{references}  %%% Uncomment this line and comment out the ``thebibliography'' section below to use the external .bib file (using bibtex) .





% \end{thebibliography}

\end{document}